\documentclass[a4paper,conference]{IEEEtran}
\usepackage{cite}
\ifCLASSINFOpdf
\usepackage[pdftex]{graphicx}
\else
\fi
\usepackage{amsmath}

\usepackage{url} 
\usepackage{multirow}
\usepackage{xcolor}
\usepackage{amsfonts}
\usepackage[utf8]{inputenc}

\usepackage{xspace}
\newcommand*{\eg}{e.g.\@\xspace}
\newcommand*{\ie}{i.e.\@\xspace}
\newcommand*{\etal}{\textit{et al.}\@\xspace}

\usepackage{threeparttable}

\begin{document}
\title{VReBERT: A Simple and Flexible Transformer for Visual Relationship Detection}

\author{\IEEEauthorblockN{Yu Cui$^{1*}$, Moshiur Farazi$^{2,1}$}
\IEEEauthorblockA{$^{1}$Research School of Computer Science, The Australian National University (ANU) \\
$^{2}$Data61--Commonwealth Scientific and Industrial Research Organisation (CSIRO)\\
Canberra, Australia
}
\texttt{u7016518@anu.edu.au}
}


%


\maketitle

\begin{abstract}
Visual Relationship Detection (VRD) impels a computer vision model to \emph{see} beyond an individual object instance and \emph{understand} how different objects in a scene are related. The traditional way of VRD is first to detect objects in an
image and then separately predict the relationship between the detected object instances. Such a disjoint approach is prone to predict redundant relationship tags (i.e., predicate) between the same object pair with similar semantic meaning, or incorrect ones that have a similar meaning to the ground truth but are semantically incorrect. To remedy this, we propose to jointly train a VRD model with visual object features and semantic relationship features. To this end, we propose VReBERT, a BERT-like transformer model for Visual Relationship Detection with a multi-stage training strategy to jointly process visual and semantic features. We show that our simple BERT-like model is able to outperform the state-of-the-art VRD models in predicate prediction. Furthermore, we show that by using the pre-trained VReBERT model, our model pushes the state-of-the-art zero-shot predicate prediction by a significant margin (+8.49 R@50 and +8.99 R@100).
\end{abstract}


%
\IEEEpeerreviewmaketitle

\section{Introduction}
Understanding how different objects are related in an image is a higher level of visual perceptions, and computer vision models tasked with this need to perform far more complex vision tasks compared to simply detecting objects or their attributes. Such a higher level of image understanding has broad application. For example, a system that understands visual relationships can aid visually impaired people by describing a scene and answering intelligent questions about it \cite{antol2015vqa, anderson2018bottom, farazi2021accuracy}. Lu et al.\cite{lu2016visual} formalized the Visual Relationship Detection (VRD) task to aid the research for addressing this challenging problem. They defined the visual relationship as a pair of localized object instances connected by a predicate, \ie, object$_1$ -- relationship tag -- object$_2$. For simplicity, in the context of this paper we refer the object$_1$ and object$_2$ as \texttt{subject} and \texttt{object} respectively, and the relationship tag as a \texttt{predicate}. Further, Lu \etal proposed the VRD dataset which contained involving $100$ object categories and $70$ predicates. The predicates include verb (\eg, \texttt{drive}, \texttt{follow}), spatial relation (\eg, \texttt{over}, \texttt{on}), comparative relations (\eg, \texttt{taller than}), and many other categories. This dataset provided an effective evaluation protocol to benchmark VRD models.

\begin{figure}[ht!]
\centering
\includegraphics[width=.48\textwidth]{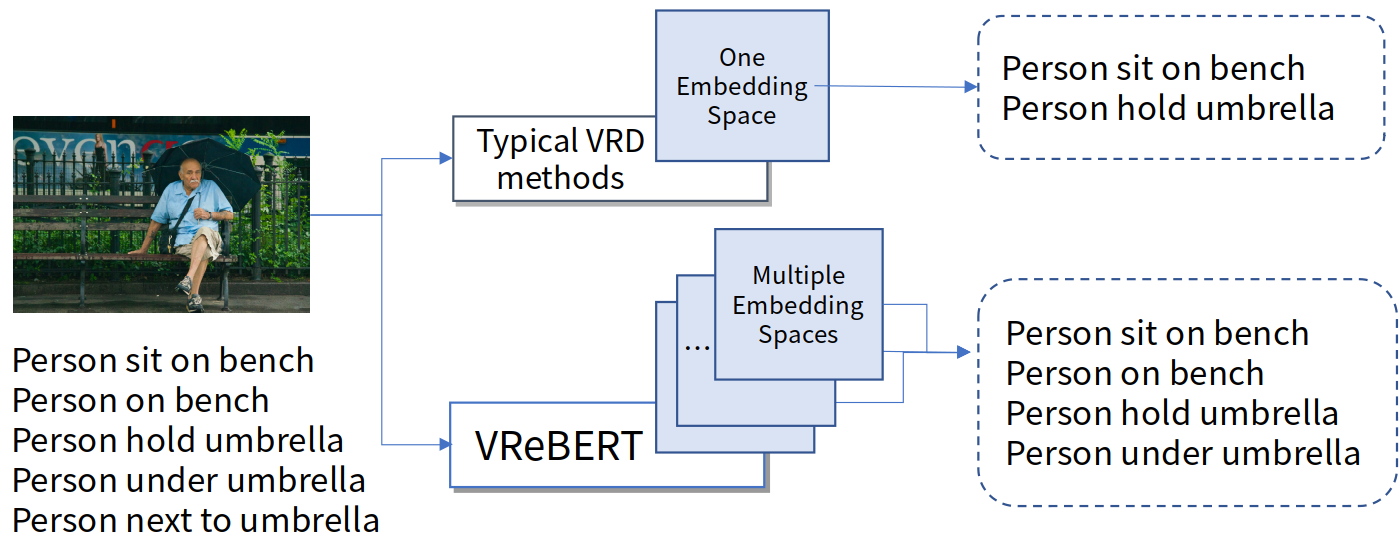}
\caption{Traditional project-based methods use only one embedding space for predicate prediction, which will limit the models’ capability of learning the context of the images. Differently, VReBERT utilizes multi-head attention to jointly learn multiple semantic meanings in different contexts.} \label{fig1}
\end{figure}

VRD models \cite{lu2016visual, zhang2017visual, younes2019block, liang2018dsr, liao2019lg, yu2015lkd} approach relationship detection by decomposing the task into two tasks. They first identify and localize two object instances as \texttt{subject} and \texttt{object} pair, and then try to predict the \texttt{predicate} between these object instances. There are a couple of fundamental challenges that an AI model will need to overcome to succeed in visual relationship detection following this approach. \textit{First,} the multi-modal aspect of the task. The objects in the images provide visual cues for a Visual Relationship Detection (VRD) model to generate both semantic (\ie, relationship tags) and visual (\ie, a localized pair of objects) output. \textit{Second,} the relationship tags (\ie, predicates) generated by a VRD model are often ambiguous as the model tries to learn visual relationships from looking and relationships between the same object pairs. It's easy for the model to forget the context and predict a synonymous relationship tag that is contextually incorrect (\eg, \texttt{holding} vs \texttt{carrying}) or fail to predict multiple relationships that may exist between the same pair of object instances (\eg, \texttt{<person-on-bicycle>} and \texttt{<person-riding-bicycle>}. Furthermore, predicates may have very different semantic meanings in different contexts, \texttt{<picture-on-wall>} and \texttt{<bottle-on-table>} share the same predicate \texttt{on} while the semantic meanings are different. To address the challenges of the multi-modal aspect of VRD and to reduce the ambiguity in predicate prediction, one would need to bridge the semantic gap between the two tasks that VRD is decomposed into. We propose to jointly process the semantic relationship tags and visual object instance features to bridge this gap. 

Recently, some transformer-based cross-modal methods achieved great success in vision and language tasks \cite{tan2019lxmert, Su2020VL-BERT, chen2020uniter}. These methods take advantage of large-scale mono-modal (\eg, vision or language) or multi-modal (\eg, vision and language) pre-training. The large-scale pre-training not only reduces the training cost but also improves the performance on a very small task-specified dataset as the training does not start from scratch. More importantly, the transformer performs well in many sequence-to-sequence problems. Inspired by the success of transformers in cross-modal vision and language tasks, in this paper we endeavour to find a sequence-to-sequence function that projects the image patch sequence to a relationship sequence. In this paper, we propose VReBERT, a simple and flexible transformer-based model designed for visual relationship detection. The flexible nature of VReBERT allows any typical object detector module (\eg, Faster R-CNN \cite{ren2015fasterrcnn}, UperNet \cite{xiao2018unified} + Swin-transformer \cite{liu2021swin}) to be used with it jointly.

Our proposed VReBERT is similar to BERT \cite{devlin-etal-2019-bert} at the conceptual level but redesigned to solve cross-modal visual relation detection. Just like BERT, and other transformer-based vision and language models VReBERT also benefits from pre-training on large-scale datasets. We show that our model achieved a new state-of-the-art performance in the visual relationship task. Furthermore, with VReBERT, we were able to improve zero-shot visual relationship detection performance significantly ($\sim$ +9 on R@50 and R@100) leveraging large-scale pre-training. This further strengthens the rigidity of our approach of jointly training the relationship detector with visual and semantic features using transformer architecture. Such strong generalization ability would help our proposed VRD model to be useful in other downstream tasks. 

The main contributions of our paper are as follows:
\begin{itemize}
    \item We propose a novel visual relationship detection framework VReBERT that can jointly learn the representation of visual and semantic contents by utilizing the extracted object location, object label, image feature, and language priors to detect relationships between an object pair in an image.
    \item We introduce a multi-stage training method for VReBERT to reduce the training cost and improve the generalization ability.
    \item We reported state-of-the-art performance with VReBERT in Visual Relationship Detection \cite{lu2016visual} task and a significant improvement in zero-shot visual relationship detection.
\end{itemize}

\begin{figure*}
\centering
\includegraphics[width=16cm]{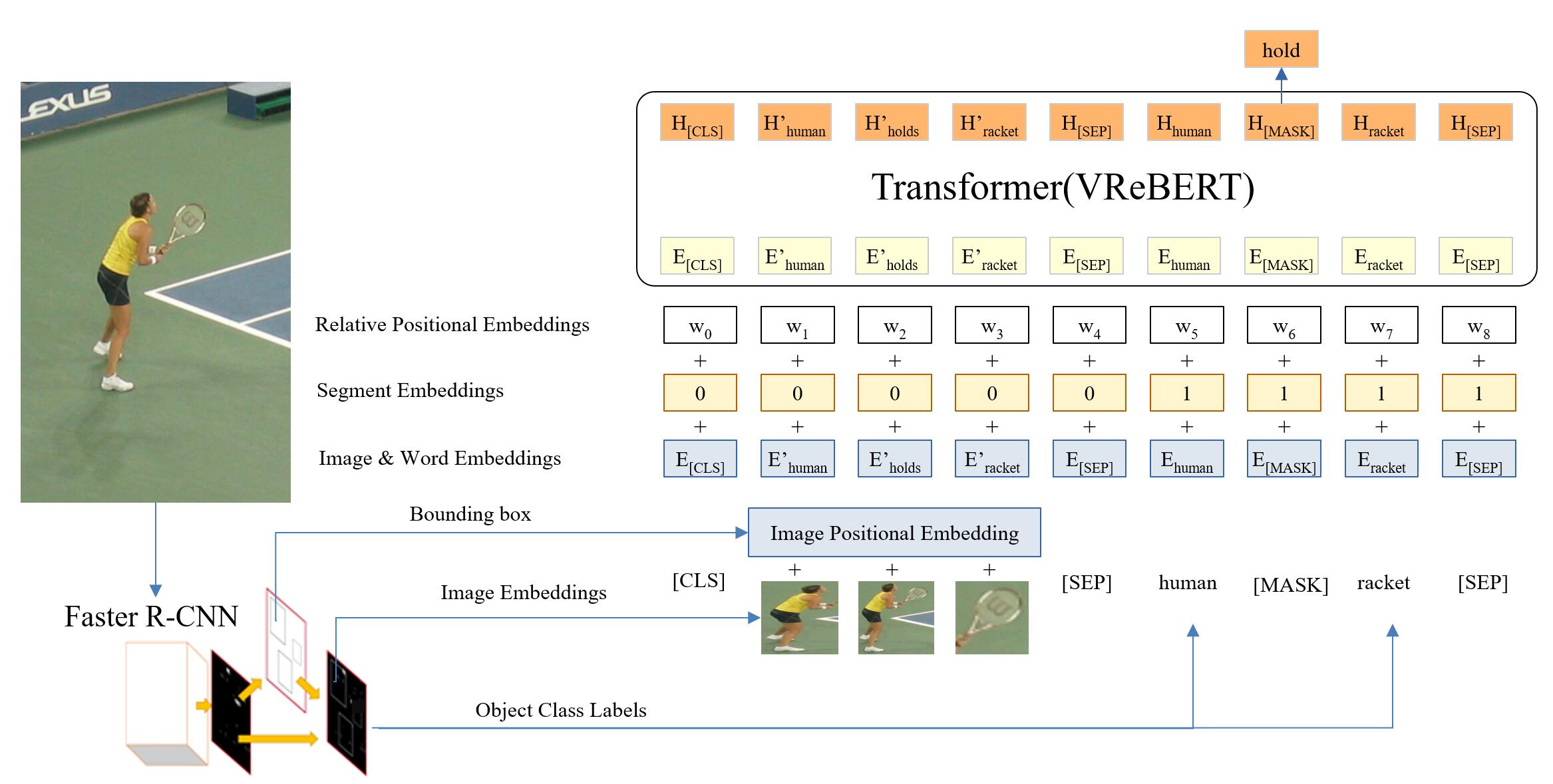}
\caption{The architecture of our proposed VReBERT model. Visual features from the image are extracted by a Faster R-CNN object detector. The visual features and their labels are combined as the input and added with segment and relative positional embedding. They are then passed to the VReBERT block that jointly learns the context of the image and language. Such joint training allows our proposed model to understand the alignment and semantic meanings and make predicate prediction in a way similar to the task of masked language modeling.}
\label{fig2}
\end{figure*}\ 

\section{Methodology}
We designed the VReBERT as a combination of modular blocks to provide greater flexibility and further development. First is the object proposal module. For this work, we use Faster-RCNN\cite{ren2015fasterrcnn} as the object detector, but as mentioned before this module can be replaced with other typical object detectors. This module takes an RGB image as input and proposes regions of interest (ROI) where it thinks there might be an object, provides a prediction of an object class for each ROI, and extracts visual features by its backbone network (\eg, ResNet\cite{he2016resnet}). VReBERT utilizes bounding boxes, class labels, and the extracted features for visual and semantic reasoning. VReBERT models the relationship prediction as a masked language model prediction task similar to a vanilla BERT \cite{devlin-etal-2019-bert}. The key differences are multi-modal inputs and redesigned embeddings for image tokens. VReBERT can jointly learn the representation of image and language to capture the interaction between the proposed objects, which allows VReBERT to solve cross-modal problems. We propose a multi-stage training strategy for the architecture mentioned above. Pre-training on the large-scale image and text dataset can help our model get a better result on the task-specified dataset. The architecture of our model is illustrated in Fig.~\ref{fig2}.

\subsection{Object Detector}
As the foundation of visual relationship detection, the object detection module sits at the beginning of the pipeline. Considering that this paper focuses on visual and semantic reasoning of the relationships, we would like the object detection module to be as simple and easy-use as possible. We employ a Faster R-CNN \cite{ren2015fasterrcnn} as the object detection module. The two main reasons for our choice are as follow. First, Faster R-CNN is fast and easy to use with lots of open-access pre-trained models which is beneficial for large-scale pre-training. Compared to other object detection frameworks, Faster R-CNN is also very competitive in terms of effectiveness. Second, Faster R-CNN \cite{ren2015fasterrcnn} uses a modular design. It contains an RPN \cite{ren2015fasterrcnn} and a Fast R-CNN \cite{girshick2015fastrcnn}. We can take advantage of this design and make the training and evaluation faster. The details will be covered in Sec.~\ref{sec:multi-stage-training}. The object detector module proposes ROIs and corresponding object category labels. ROIs include the bounding box and the representation extracted from the image. The labels come with classification probabilities. We extracted the object bounding box, region representation, category label, and category confidence for relationship reasoning.

\subsection{VReBERT}
Many of the previous works \cite{lu2016visual,zhang2017visual,qi2019attentive,nagaraja2016context} tried to project the subject-predicate-object sequence to the semantic embedding space to understand the visual relationships, which can be generically expressed as: 
\begin{equation}
S_{text} =  \textrm{F}_{seq2seq}(S_{image})
\end{equation}

Furthermore, Yun \etal \cite{yun2020universal} mathematically showed that transformers are universal approximators of sequence-to-sequence functions. Inspired by this, we want to find a sequence-to-sequence function that projects the image feature sequence to a relationship sequence. As mentioned earlier, previous works struggled with two challenges: one predicate with multiple semantic meanings and multiple correct predicates between one subject-object pair. To conquer the first challenge our method has to be able to model the polysemy (\ie, a word or phrase having multiple related meanings) and understand the different meanings of a predicate. For the second challenge, we need to utilize both visual and linguistic information. Our model needs to have an excellent multi-modal reasoning capability. The multi-head attention mechanism proposed with the transformer is designed to jointly understand different representation sub-spaces, which can efficiently deal with the polysemy. Different attention heads can learn the different semantic meanings of a predicate. When looking at a predicate, the transformer can leverage all attention heads and consider all possible meanings. Inspired by \cite{devlin-etal-2019-bert, qi2020imagebert, li2019visualbert}, we propose VReBERT, a BERT-like transformer model for visual and semantic reasoning.

\subsection{Multi-modal Embedding}
We follow similar token nomenclature of vanilla BERT and use special tokens like \texttt{[CLS]} and \texttt{[SEP]} as the task indicator and sentence divider. The word embedding is kept basically the same, while we redesign the image embedding for visual relationship reasoning.

\subsubsection{Word embedding}
The sentence input contains the word-based representation of object and subject. Then we use WordPiece \cite{wu2016nmt} method to tokenize the word-based object labels from the object detection module into sub-word tokens. We add a special token \texttt{[MASK]} between the token of subject and object for masked language modeling. After we get the token representation of the relationship sequence, we add the segmentation embedding and position embedding to the tokens like the vanilla BERT. 

One thing worth noting is that we utilize a relative positional embedding method instead of the absolute position embedding used in BERT. The attention mechanism cannot comprehend the sequence order from the word tokens. So \cite{vaswani2017attention} and \cite{devlin-etal-2019-bert} use a fixed sinusoid function to encode the sequence order of the tokens. The absolute positional embedding does not perform well in some tasks (\eg, question answering \cite{huang2020improvedrp}). Furthermore, Yun \etal \cite{yun2020universal} showed that trainable positional encoding is important for transformer-based models to be a universal approximator of a sequence-to-sequence function. Recently, transformer-based methods \cite{shaw2018rp, yang2019xlnet, dai2019transformerxl} adopt relative position encoding as trainable tokens. 
In our paper, we employ a relative positional encoding method similar to \cite{huang2020improvedrp} as follow:
\begin{equation}
    \mathbf{a}_{ij} = \mathbf{w}_{j-i}
\end{equation} 
where $\mathbf{a}_{ij}$ represents the relative position embedding of position $j$ with respect to position $i$, $\mathbf{w}_{j-i}$ is a trainable vector. The position embedding will interact with both the queries and the keys during self-attention. The modified scale attention $Attn_{ij}$ can be represented as:
\small{
\begin{equation}
Attn_{ij} = \frac{(\mathbf{x}_{i}\mathbf{W}^Q)\cdot(\mathbf{x}_{j}\mathbf{W}^K) + \mathbf{W}^Q \cdot \mathbf{a}_{ij} + \mathbf{W}^K \cdot \mathbf{a}_{ij}}{\sqrt{d_z}}
\end{equation}
}
We expect this trainable relative positional embedding to learn unique position tokens to capture the impact of the position on attention. More specifically, besides the sequence order, we want it to learn the alignment of image and text.

\subsubsection{Image Embedding}
\label{sec:image-embedding}
Image embedding consists of visual features and image positional embedding.

\noindent \textbf{Visual Features:} The image is represented as a combination of visual features and positional embeddings. The visual features of the bounding box containing the $subject$, $object$, and $predicate$ (\ie, the bounding box containing both subject and object) are represented as $\mathbf{i}_{sub}$, $\mathbf{i}_{pred}$, $\textbf{i}_{obj}$ respectively which are all $\in \mathbb{R}^D$.

\noindent \textbf{Image positional embedding:} Further to the visual features, the spatial location of the relationship triplet is parameterized by normalized bounding box locations. Each bounding box is represented  by $[X_{min}, Y_{min}, X_{max}, Y_{max}]$ and is normalized by the corresponding height $H$, or weight $W$ dimension of the image. We further add another spatial area normalization term for the bounding boxes to normalize the bounding box area with respect to the total image area. The resultant 5-dimension image position embedding vector is projected to $D$-dimension using a fully connected layer $\mathbf{P}$. The image position embedding can be represented as
\begin{equation}
\small{
\begin{split}
\mathbf{p}_i=\textrm{P}(\frac{X_{min}}{W},\frac{Y_{min}}{H},\frac{X_{max}}{W},\frac{Y_{max}}{H},\\
\frac{(X_{max}-X_{min})*(Y_{max}-Y_{min})}{W*H})
\end{split}}
\end{equation}

\subsubsection{Special embedding} Additional two kinds of special embedding are used to train our model in a BERT-like manner. We use the same segment embedding to delineate between image embedding and relationship sequence by setting the former to $0$ and the latter to $1$.  We also use the relative position embedding mentioned earlier for image sequence position embedding to force the relationship patches to have the triplet order \texttt{<sub-pred-obj>}. This relative positional embedding is learned by the model by understanding the alignment between the image tokens and the word tokens.

\subsection{Relationship likelihood}
For relationship detection, we need to consider both object likelihood and predicate likelihood. We use the following likelihood function for relationship:
\begin{equation}
\begin{split}
P_{i,j,k}(pred, obj, sub) =
P_i(pred\ |\ obj,\ sub) P_j(obj) P_k(sub)\hfill 
\end{split}
\end{equation}
where $P_j(obj)$ and $P_k(sub)$ are the likelihood from the object detection module for object and subject being in a specific category and $P_i(pred\ |\ obj, sub)$ is the likelihood from the VReBERT, which indicates the predicate belongs to category $i$.

\subsection{Multi-stage Training Strategy}
\label{sec:multi-stage-training}
Having a flexible and lightweight training strategy, the generalization capability of transformer-based models increase significantly. With the large-scale pre-training and task-specified fine-tuning, BERT-like models \cite{devlin-etal-2019-bert,qi2020imagebert,li2019visualbert} achieved great success. Since our model jointly learns the representation of image and language, our pre-training stage is more complex than only training on language or image. We implement the following multi-stage training strategy to reduce the training and evaluation cost and to be able to run some task-specified benchmarks to gauge the performance of the model undergoing the full training. This strategy is designed for the VReBERT model with a typical object detector (Faster R-CNN in our case).

\begin{itemize}
    \item[S1] \textbf{Pre-training on large-scale datasets:} The object detector used pre-trained weights learned from PASCAL VOC 2012, and  VReBERT is initialized from a pre-trained BERT$_{base}$ \cite{devlin-etal-2019-bert} model.
    \item[S2] \textbf{Separate training of object detector and the transformer core:} The object detector is re-trained on the target Visual Relationship Dataset \cite{lu2016visual}. The BERT part of our model is separately trained on masked relationship prediction where the input to the models is \texttt{<sub-[MASK]-obj>} and the model tries to predict the relationship only from language cues without looking at the image. Such separate mono-modal training allows the models to `learn' task-specific capabilities which they can build during the multi-modal training. 
    \item[S3] \textbf{Training VReBERT on the multi-modal VRD dataset:} After the training on the mono-modal task, VReBERT can handle linguistic reasoning and object detection. This stage aims to combine these two abilities for visual relationship detection. Now, the VReBERT model trained image representations (\ie, visual features, image positional embedding as discussed in \ref{sec:image-embedding}) extracted using the object detector, word embeddings similar to that of the masked relationship prediction stage, segment and relative positional embedding. One thing worth noting is that the predicate benchmark in \cite{lu2016visual} uses the same input and output setup. Thus, we can evaluate VReBERT’s ability to predict the predicate after this stage of training, and we report our model performance on predicate prediction from this stage for fair comparison.
    \item[S4] \textbf{Fine-tuning:} We fine-tune the region proposal network (RPN) for the object detector, to make the pipeline end-to-end. This stage helps the VReBERT to get better object proposals for predicate prediction with a combined loss function during fine-tuning our model on the Visual Relationship Dataset.
\end{itemize}

\section{Experiments}
\subsection{Implementation details}
Our implementation is on the PyTorch platform. We employ a Faster R-CNN with ResNet101 \cite{he2016resnet} pre-trained on ImageNet as the backbone. The image feature output is set to $2048$-dimensional. The image feature is projected to the same dimension as the VReBERT linguistic embeddings. The VReBERT is configured with $768$ hidden dimensions, 12 self-attention heads, and 12 hidden layers. The dropout probability is set to 0.1 to improve the generalization capability. We employ the AdamW optimizer with an initial learning rate of $1\times 10^{-5}$ and set $\beta_1=0.9$, $\beta_2=0.999$.

\subsection{Dataset}
In this paper, we train and evaluate our model on the VRD dataset \cite{lu2016visual}. This dataset contains $5000$ images with $100$ object categories and 70 predicate categories. It also has $37,993$ relationships in total, $6672$ relationship types are collected resulting in $7.6$ relationships per image. Each object category is related to $24.25$ predicates on average. We follow the dataset split in \cite{lu2016visual}. We use $4000$ images to train our model and $1000$ images to test. The test set contains $1877$ relationships that never appear in the training set. The unseen relationships can help us evaluate the performance gain from large-scale pre-trained and the model's generalization ability. Even though VRD is a relatively small dataset, we choose this to benchmark our model can be trained with such dataset size leveraging large-scale pre-training.

\subsection{Evaluations metrics}
Following \cite{lu2016visual}, we report Recall@50 (R@50) and Recall@100 (R@100) as our evaluation metric for image relationship detection. Recall@N indicates how many true relationships are detected by the model in the top-N results. Recall is defined as a weighted representation of True Positive (TP), considering False Negatives (FN), and is expressed as:
\begin{equation}
Recall = \frac{TP}{TP + FN}
\end{equation}

\subsection{Predicate Prediction}
Predicate prediction is the task of predicting the relationship tag (\ie, \texttt{predicate}) based on the ground truth object position and label. For this task, we want to test the predicate reasoning ability of our method. So, we use the ground truth visual information to avoid the performance impact from the object detection module. This task can directly compare VReBERT with the previous visual relationship detection methods on the predicate prediction ability. With the multi-stage training strategy, we can analyze the predicate prediction performance after the third stage of training as previously discussed in Sec.~\ref{sec:multi-stage-training}. By doing so, even without fully training the entire model, we can have a general idea about the performance of our model. Considering that predicate prediction is the key to visual relationship detection, the experiment will mainly focus on this task.

\subsubsection{Comparable methods}
We compare our method with the following comparable methods in the literature utilizing a diverse and comprehensive set of approaches. First, with the VRD \cite{lu2016visual} where the projection-based word2vector method was proposed together with the Visual Relationship Detection problem. The LKD \cite{yu2015lkd} used knowledge distillation \cite{hinton2015distilling} utilizing both student and teacher networks to distill internal and external linguistic knowledge for visual relationship detection. Further, DR-Net\cite{dai2017dr} is an RNN-based joint recognition method utilizing both image feature and linguistic information. Furthermore, DSR\cite{liang2018dsr} proposed a visual relationship detection method based on deep structure ranking. Finally with BLOCK\cite{younes2019block} fusion model used linear superdiagonal fusion to combine visual and semantic features for visual relationship detection.

\subsubsection{Experimental setup}
\label{sec:experimental_setup}
We perform ablation with different combinations of our VReBERT models report their performance on Tab.~\ref{tab1}. The different combinations are 
\begin{itemize}
    \item[ES1] \textbf{BERT from scratch:} In this setting, we only employ a vanilla BERT. We train this model from scratch using the relationship labels from the VRD dataset to show the performance of the pure language model.
    \item[ES2] \textbf{Pre-trained BERT:} This model initialized from a BERT model pre-trained on large-scale datasets. We fine-tuned this model on the VRD dataset.
    \item[ES3] \textbf{VReBERT with pre-trained visual features}: We use image features from a pre-trained Fast R-CNN (After S2 pre-training in Sec.~\ref{sec:multi-stage-training}) features for predicate reasoning. And we use absolute sequence position encoding.
    \item[ES4] \textbf{VReBERT with fine-tuned visual features:} We fine-tune the Fast R-CNN with VReBERT and use the fine-tuned image feature for prediction. And we use absolute sequence position encoding.
    \item[ES5] \textbf{VReBERT with image positional embedding:} We add image positional embedding (IP) to compare the performance gain from image positional embedding.
    \item[ES6] \textbf{VReBERT full model:} We replace the absolute sequence positional encoding with relative positional encoding.
\end{itemize}

\begin{table}
\caption{Predicate Prediction Performances of Different Methods.}\label{tab1}
\centering
\renewcommand{\arraystretch}{1.1}
\begin{tabular}{|p{4cm}|p{1.5cm}|p{1.5cm}|}
\hline
Method  & R@50  (\%) & R@100 (\%)\\
\hline
VRD-L only \cite{lu2016visual}& 48.64 &  61.95\\
VRD-Full \cite{lu2016visual} & 70.97 &  84.34\\
LKD \cite{yu2015lkd}& 85.64 & 94.65\\
DR-Net \cite{dai2017dr}& 80.78 & 81.90\\
DSR \cite{liang2018dsr}& 86.01 & 93.18\\
BLOCK \cite{younes2019block}& 86.58 & 92.58\\

\hline
BERT from scratch & 54.50  & 71.29\\
Pre-trained BERT  & 62.35  & 79.34\\

\hline
VReBERT with PF + AP & 61.40  & 85.11\\
VReBERT with FF + AP & 83.24  & 92.77\\
VReBERT with FF + AP + IP & 88.48  & 94.90\\
VReBERT with FF + RP + IP& $\mathbf{88.57}$  & $\mathbf{95.26}$\\

\hline
Improvement over SOTA & \textbf{2.93 $\uparrow$} & \textbf{0.61 $\uparrow$}\\
\hline
\end{tabular}
\begin{tablenotes}
\scriptsize{
\item The first set of models are from literature, the middle set is our blind VReBERT (no visual input), and the last set of models is multimodal VReBERT.
\item `PF' is the visual feature generated by the pre-trained Faster R-CNN. 
\item `FF' is the visual feature generated by the fine-tuned Faster R-CNN. 
\item `AP' is absolute positional embedding using sinusoid”. 
\item `RP' is the trainable relative positional embedding. 
\item `IP' is image positional embedding.}
\end{tablenotes}
\end{table}

\subsection{Result}
\begin{figure*}
\centering
\includegraphics[width=10cm]{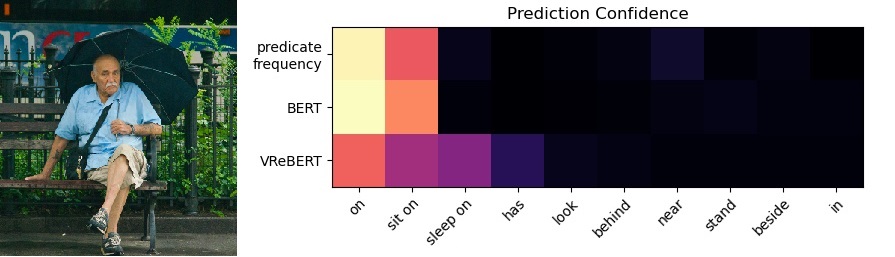}
\caption{Predicate Frequency, BERT, and VReBERT Prediction Confidence. The Language-only BERT model distributes its attention only based on the predicate frequency, while the multi-modal VReBERT leverages the image feature to propose visually reasonable predicates.} \label{fig8}
\end{figure*}

\textit{How effective are pure language models in VRD?}\\
We start with a vanilla BERT to show how a pure language model performs in this visual-language task. When comparing the BERT model with the VRD-L model \cite{lu2016visual}, the transformer-based BERT model is better than the projection-based model \cite{lu2016visual} for language modeling. As we mentioned in the dataset section, the test set contains 1887 unseen relationships. Simply training on the VRD training set may cause the model to struggle with unseen relationship prediction. In experimental setting 2 (\ie, ES2 in Sec.~\ref{sec:multi-stage-training}), we initialized VReBERT with a pre-trained BERT-base model. The result shows pre-trained BERT model achieved a better result. However, even with a large-scale pre-training, the language model can only get $62\%$ in R@50. And during the experiment, we notice that the BERT model always predicts a predicate with the higher frequency, as shown in the first two rows of (Fig. \ref{fig8}), where the subject-object pair is set to \texttt{person-bench}. The BERT prediction is basically the same as the frequency of occurrence. It does make sense considering the statistical property of the language models. This fact indicates language models have difficulty in predicting the predicate without the visual information.

\textit{Can VReBERT leverage image features for relationship reasoning?}\\
As we mentioned early, pure language does not perform well in visual relationship detection. Our model with fine-tuned visual appearance (ES4) achieved a much higher recall rate than the BERT model proving that the visual appearance does help with the relationship reasoning. In ES5, we add the image positional embedding to the image appearance. This embedding contains the location and size of the object/subject. After the image positional embedding is added, we observe a very consistent recall improvement over ES4. To analyze the impact of the image features, we show a specific example of the relationship reasoning with image features in the third row in (Fig. \ref{fig8}). We notice that even though the predicate \texttt{sleep on} never appeared between person and bench. VReBERT still has quite high confidence which prediction is reasonable both in terms of spatial relation and appearance (\eg, a person can sleep on a bench). With the support of the experiment results and the mentioned case, we believe VReBERT can leverage image features for relationship reasoning which complement each other.
\begin{figure}
\centering
\includegraphics[width=9cm]{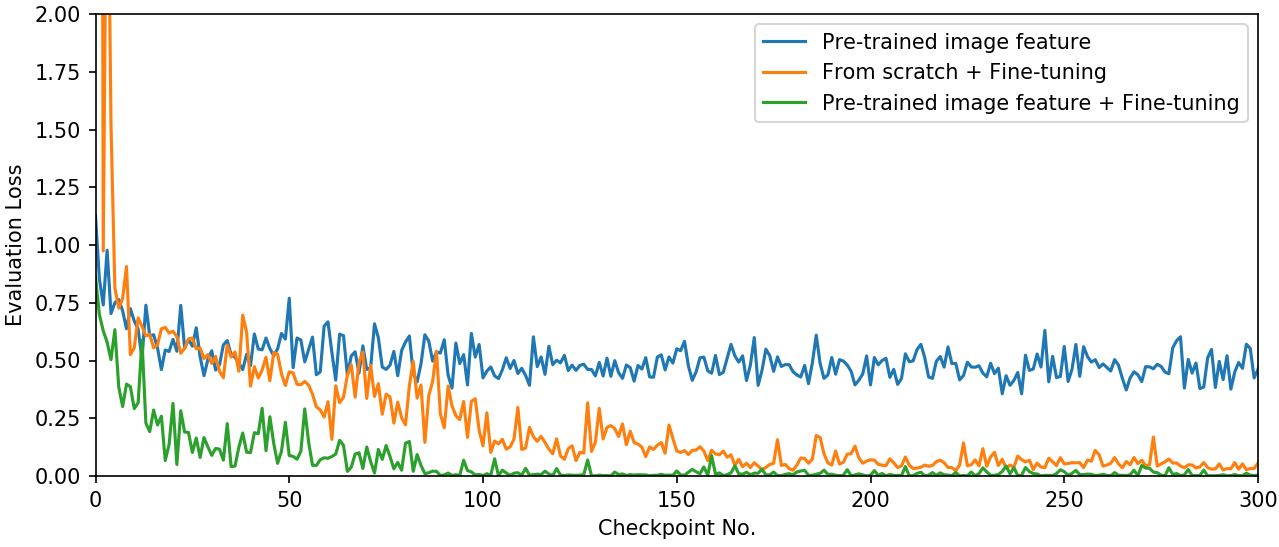}
\caption{Evaluation Losses of Different Training Settings. Initializing from a large-scale pre-trained visual model can dramatically reduce the training cost. With end-to-end fine-tuning, VReBERT achieved noticeable improvement than using the pre-trained image feature.} \label{fig9}
\end{figure}

\begin{figure}
\centering
\includegraphics[width=6cm]{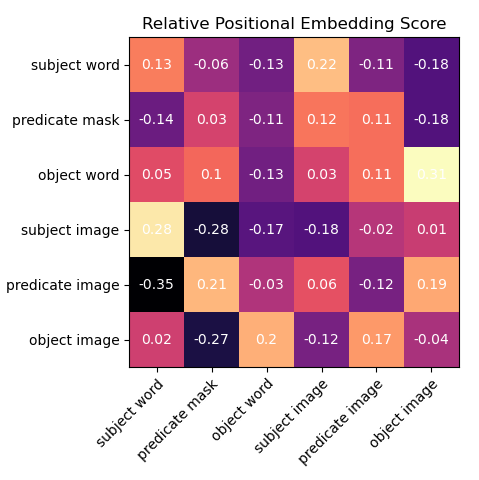}
\caption{Relative Sequence Positional Embedding Score. This figure shows that the trainable sequence positional embedding properly learned the alignment of the image and text. For example. The embedding shows a significantly higher value between the predicate mask and the predicate image token, which indicates the predicate mask is highly related to the predicate image token in sequence location.} \label{fig10}
\end{figure}

\textit{Is end-to-end fine-tuning necessary for VRD?}\\
ES3 uses a Fast R-CNN model pre-trained on the VRD dataset (after S2 training as described in Sec.~\ref{sec:multi-stage-training}). We pass a pre-trained image feature to the BERT model just like \cite{lu2016visual}. We did not observe a significant improvement over the pure language model. In ES4, we fine-tune VReBERT and Fast R-CNN end-to-end on the ground truth ROIs and relationship labels (at S3 of multi-stage training) as see that VReBERT with fine-tuned image features obtained a huge improvement (R@50 from 61.40\% to 83.24\%, R@100 from 85.11\% to 92.77\%) over VReBERT with the pre-trained image features. To intuitively show the impact of fine-tuning, we analyzed the evaluation loss of different pre-training and fine-tuning settings. We randomly choose 400 images as the evaluation set. We train our model on the rest of the training set and evaluate it on the evaluation set. We show in Fig. ~\ref{fig9}, that the setting with pre-trained image feature without fine-tuning has much higher and unstable evaluation loss than the setting with fine-tuned features. We argue that BERT-like models may have trouble utilizing the pre-trained image feature extracted for multi-layer perceptron (MLP). Thus, end-to-end fine-tuning is necessary for improving the recall rate. Furthermore, the pre-training dramatically accelerates the training progress. 
\textit{Are relative sequence positional embeddings useful?}\\
We utilize relative sequence positional embedding for mainly two reasons. First, the learnable positional embedding is necessary for our model to be an approximator of universal sequence-to-sequence functions. Second, the alignment of the image and language token could be improved by it. The second reason is quite important for multi-modal problems like visual relationship reasoning. With a good alignment, the self-attention mechanism can better understand the dependency of the tokens. In our experiment, we see that ES6 has a visible but minor recall gain over ES5. It is hard to confirm that relative sequence positional embedding will always lead to a performance gain based only on the current experimental results. 
We try to visualize it in Fig.~\ref{fig10}, where we plot a graph of the relative sequence positional embedding score. This score will be added directly to the scaled attention score which can in-turn impact VReBERT’s understanding of dependency. A higher score means higher dependency on the token. From the figure, we can find out the embedding value of corresponding text and image token is much higher than other tokens, which indicates the relative sequence positional embedding properly learned the alignment of the image and language.  This further adds to the importance of such learned positional embedding.

Overall, our full model outperforms the state-of-the-art result \cite{yu2015lkd} in the predicate prediction task by $2.93\%$ in R@50 and $0.61\%$ in R@100.

\subsection{Zero-shot Predicate Prediction}
Further to traditional Predicate Prediction, we evaluate VReBERT on challenging zero-shot predicate prediction. The task is to predict the predicate category based on the ground truth object position and label in a zero-shot setting. For this task, we test our model only on the zero-shot test set which contains $1877$ relationships that never appeared in the training data. We design this test to evaluate our model's generalization capability.

We mainly focus on the performance impact of pre-training on the large-scale dataset. To compare with our VReBERT fully trained using the multi-stage training strategy, we trained a VReBERT model from scratch without pre-training on a large-scale language dataset (S1 in Sec.~\ref{sec:multi-stage-training}). We also compare VReBERT with other comparable methods (e.g., VRD \cite{lu2016visual}, LKD \cite{yu2015lkd}) in the same zero-shot setting.

We show in Tab.~\ref{tab2}, that our full model achieves $65.30\%$ in R@50 and $85.41\%$ in R@100. It is a significant improvement from the state-of-the-art models. VTransE aimed to find a proper embedding for object and predicate. When optimizing the embedding space, it is hard to make it smooth and generalize well on the unseen dataset. Later works like \cite{yu2015lkd} utilized both internal and external knowledge for visual relation detection. We think the performance boost is due to the large-scale pre-training that helps bridge the semantic gap between the relationship tags and contextual visual relationship. By doing so the model is able to predict may unseen relation (\eg, \texttt{<person-sleep on-bench}) even though it has only seen examples of \texttt{<person-on-bench>}. We hope our findings can help further research how to better utilize such pre-training for other zero-shot multimodal tasks. 

\begin{table}
\caption{Zero-shot Predicate Prediction Performances}\label{tab2}
\renewcommand{\arraystretch}{1.1}
\centering
\begin{tabular}{|p{3cm}|p{2cm}|p{2cm}|}
\hline
Method  & R@50  (\%) & R@100 (\%)\\
\hline
VRD-Full \cite{lu2016visual} &  29.77 &   50.04\\
LKD \cite{yu2015lkd}& 56.81 & 76.42\\
\hline
VReBERT from scratch & 52.06  & 81.75\\
Pre-trained VReBERT & $\mathbf{65.30}$  & $\mathbf{85.41}$\\
\hline
Improvement over SOTA                                    & \textbf{8.49 $\uparrow$} & \textbf{8.99 $\uparrow$}\\

\hline
\end{tabular}
\end{table}

\section{Conclusion}

We proposed a simple and flexible framework for visual relationship detection. Our model leverages multi-head self-attention to conquer two challenges faced by traditional visual relationship detection methods, which are polysemy understanding and context-aware prediction of multiple predicates in different categories. By redesigning the embedding layer of the BERT model, we fully take advantage of visual and linguistic features to predict the relationship. The experimental results on Visual Relationship Dataset show that our framework is in line with the best visual relationship detection methods. We show that our proposed VReBERT model is able to achieve state-of-the-art performance in the visual relationship detection tasks and significantly improve zero-shot predicate prediction. The experiment on the zero-shot test set proved that the multi-stage training strategy proposed in this paper gives VReBERT good generalization ability.






\bibliographystyle{IEEEtran}
\bibliography{bibliography.bib}


\end{document}